\documentclass{article}
\usepackage{ijcai20}

\usepackage{times}
\usepackage{soul}
\usepackage{url}
\usepackage[hidelinks]{hyperref}
\usepackage[utf8]{inputenc}
\usepackage[small]{caption}
\usepackage{graphicx}
\usepackage{amsmath}
\usepackage{amsthm}
\usepackage{booktabs}
\usepackage{algorithm}
\usepackage{algorithmic}
\usepackage{amsmath,verbatimbox}

\title{Reinforcement Learning with External Knowledge \\ by using Logical Neural Networks}

\author{
Daiki Kimura
\and
Subhajit Chaudhury
\and
Akifumi Wachi
\and
Ryosuke Kohita
\and \\ \vspace{3pt}
Asim Munawar
\and
Michiaki Tatsubori
\and
Alexander Gray
\affiliations
IBM Research AI
\emails
daiki@jp.ibm.com
}

\begin{document}

\maketitle

\begin{abstract}
Conventional deep reinforcement learning methods are sample-inefficient and usually require a large number of training trials before convergence. Since such methods operate on an unconstrained action set, they can lead to useless actions. A recent neuro-symbolic framework called the Logical Neural Networks (LNNs) can simultaneously provide key-properties of both neural networks and symbolic logic. The LNNs functions as an end-to-end differentiable network that minimizes a novel contradiction loss to learn interpretable rules. In this paper, we utilize LNNs to define an inference graph using basic logical operations, such as AND and NOT, for faster convergence in reinforcement learning. Specifically, we propose an integrated method that enables model-free reinforcement learning from external knowledge sources in an LNNs-based logical constrained framework such as action shielding and guide. Our results empirically demonstrate that our method converges faster compared to a model-free reinforcement learning method that doesn't have such logical constraints.
\end{abstract}

\section{Introduction}

Deep reinforcement learning methods have been successfully applied to many applications, particularly computer game, text-based game, and robot control 
applications~\cite{dqn,narasimhan2015language,alphazero,kimura2018daqn,curiosity,yuan2018counting,marioirl,vigan}. 
%applications~\cite{dqn,narasimhan2015language,curiosity,yuan2018counting,marioirl}.
Such methods require a large number of training trials for converging to an optimal action policy. By default, due to a lack of external constraints, they cannot avoid unsafe or useless actions. If an agent receives the proper action list (recommendations for action), it can reduce the number of training trials. We believe we can prepare such an action list from external action constraints pertaining to the environment. Another option is safe reinforcement learning~\cite{garcia2015comprehensive}, which can avoid taking unsafe and useless actions via the action constraints. 

To define such action constrains, there are various techniques in reinforcement learning that use symbolic logical functions or graphs~\cite{hasanbeig2018logically,hasanbeig2020cautious}. However, these techniques require all rules to be set manually, which is time-consuming. A recent neuro-symbolic framework called the Logical Neural Networks~(LNNs)~\cite{riegel2020logical} simultaneously provides key-properties of both neural networks~(learning) and symbolic logic~(reasoning). It can train the constraints and rules with logical functions in the neural networks, and since every neuron in the LNNs has a component for a formula of weighted real-valued logics, it can calculate the probability and contradiction loss for each of the propositions. At the same time, trained LNNs follow symbolic rules, which means they yield a highly interpretable disentangled representation. In this paper, we define the external knowledge for the reinforcement learning within this LNNs structure, and then leverage the knowledge in the LNNs.

We propose an integrated reinforcement learning method with external knowledge for action shielding~(avoiding useless action) and guiding~(giving an action recommendation) that is defined in the LNNs. The performance of our proposed method is experimentally compared with a baseline method that doesn't use external knowledge. Our main contributions are (1) the proposal of an integrated method that uses logical guides for fast training and safe reinforcement learning via the trainable logical network and (2) the experimental demonstration for the effectiveness of external knowledge in reinforcement learning by using LNNs.

\section{Proposed Method}

\begin{figure}[t]
\begin{center}
\includegraphics[width=8.3cm]{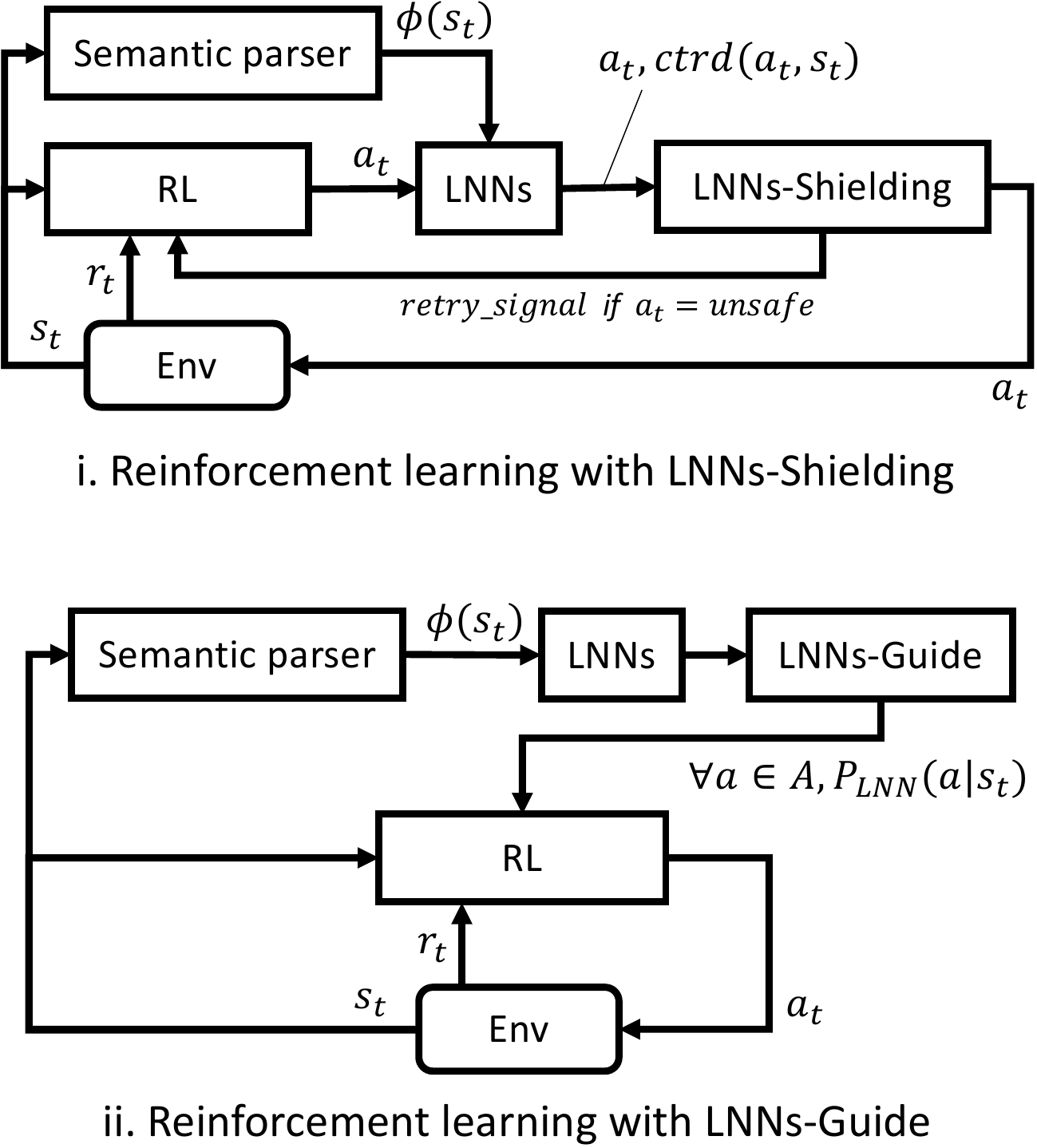}
\end{center}
\caption{Architecture of reinforcement learning with proposed methods. }
\label{fig:arch}
\end{figure}

In this paper, we proposed two methods: safe reinforcement learning by LNNs~(LNNs-Shielding) and action-guided reinforcement learning by LNNs~(LNNs-Guide). Figure~\ref{fig:arch} shows the architectures of reinforcement learning with these methods, we provide a detailed explanation of each in the following subsections. Since we apply these methods in a text-based game environment that is specifically Coin-collector game of TextWorld~\cite{cote2018textworld}, our explanations include some examples from on the game. In Coin-collector game, the agent needs to find and take a coin in series of connected rooms and take the coin. Hence, the agent needs to travel though various rooms while seeking the coin.

\subsection{LNNs-Shielding}

LNNs-Shielding avoids unsafe or useless actions that are defined by logical conditions in the LNNs for reinforcement learning. For example, the agent had better not to take the action of ``\texttt{go to west room}'', if it has already visited the west room. Action shielding needs to be represented in logical operations, hence the logical function for this example is ``\texttt{visited west room}''~$\land$ ``\texttt{found west room}''~$\Rightarrow$~$\lnot$~``\texttt{go to west room}'' ($\land$: AND operator, $\Rightarrow$: IMPLY operator, $\lnot$: NOT operator). This shielding means that even if the agent has found the west room, it will not go to the west room if the west room has already been visited. In the LNNs, the agent first checks the current state values, then if it has already visited the west room, the agent set a true value for proposition of ``\texttt{visited west room}'', that means ``\texttt{visited west room}'' = \textit{true}. If the agent find exit for the west room, it also set a true value for ``\texttt{found west room}''. Then the LNNs have a logical function for this proposition, which is ``\texttt{found west room}''~$\Rightarrow$~``\texttt{go to west room}''.

At same time, the agent inputs current state values to RL method to obtain the action. If the RL method outputs ``\texttt{go to west room}'' as a selected action, the agent set a \textit{true} value for a proposition of ``\texttt{go to west room}'' action in LNNs. If the agent visited and found the west room~(``\texttt{visited west room}'' = ``\texttt{found west room}'' = \textit{true}) and RL method outputs the ``\texttt{go to west room}'' as a selected action, the ``\texttt{go to west room}'' proposition will observe a contradiction in the proposition. Because the proposition was set \textit{true} value from RL method by the selected action candidate, it was also set \textit{false} from logical function (``\texttt{found west room}''~$\Rightarrow$~``\texttt{go to west room}'').
We assume such action restrictions will help lead to faster convergence in reinforcement learning.

Let~$\phi(s_t)$ be a logical propositional state input for current state~$s_t$ from a semantic parser algorithm. We can obtain these logical state values from raw text descriptions via the semantic parser such that the statement ``\texttt{found west room}'' is \textit{true} when the state description has ``\texttt{There is an unguarded exit to the west.}''~\footnote{The semantic parser is not our focus in this paper.}. Let~$\langle lower_{n_i,s_t}, upper_{n_i,s_t} \rangle=LNN(n_i|\phi(s_t))$ be \textit{lower} and \textit{upper} bound values from the LNNs for a node~$n_i$ and input~$\phi(s_t)$. All neurons in the LNNs return pairs of values in the range~$[0, 1]$ representing \textit{lower} and \textit{upper} bounds on the truth values of their corresponding subformulae and propositions~\cite{riegel2020logical}. These values are updated using an inference function from given propositional inputs. Note that the weight and bias values in the connections are updated during the back-propagation operations. Normally, the \textit{upper} bound is higher than the \textit{lower} bound. However, if a neuron is observed to be contradicting the logical rules, the \textit{lower} bound will be higher than the \textit{upper} bound. Therefore, the contradiction value~$ctrd(a_t, s_t) $ for the node~$n_i$ is defined as

\begin{align}
ctrd(a_t,s_t) = \sum_{n_j \in {\rm to}(a_t)} {\rm max} (0, lower_{n_j,s_t}-upper_{n_j,s_t}),
\end{align}

\noindent where~$a_t$ represents a node for an action value at~$t$ time step from the model-free reinforcement learning method, and ${\rm to}(a_t)$ is all nodes connected to node~$a_t$. 

In reinforcement learning, the agent calculates this contradiction value from a given action~$a_t$ and state~$s_t$ from the model-free LSTM-DQN++~\cite{yuan2018counting} reinforcement learning method. The proposed LNNs-Shielding distinguishes whether the given action is \textit{safe}~(useful) or \textit{unsafe}~(useless) from the inference result in the LNNs. The action~$a_t$ will be discriminated as a \textit{safe} action if the contradiction value~$ctrd(a_t,s_t)$ is~$\alpha$ or higher. The action~$a_t$ will be discriminated as an \textit{unsafe} action if the condition value is less than~$\alpha$.  When the action~$a_t$ is a \textit{safe} action, the agent executes action~$a_t$. Alternatively, when it is an \textit{unsafe} action, the LNNs-Shielding returns the action~$a_t$ to the reinforcement learning module, and the module then calculates the next candidate for proper action. Note that the contradiction value of this next candidate will also be checked by the LNNs-Shielding. The action policy training from the reward signal is then performed by the reinforcement learning method.

\subsection{LNNs-Guide}

LNNs-Guide recommends the suitable actions that are defined by the logical functions in LNNs for reinforcement learning. For example, if the LNNs are trained similar rules to those in the previous LNNs-Shielding example, the LNNs-Guide can give a negative recommendation (similar to the shielding) for the visited room. At the same time, the LNNs-Guide can recommend taking a ``\texttt{go to west room}'' action when the agent has found the west room, which is a positive recommendation. For this example, we implement this rule on various logical operations such as ``\texttt{found west room}'' $\Rightarrow$ ``\texttt{go to west room}''. LNNs-Guide, therefore, can additionally give positive recommendations compared to LNNs-Shielding. We assume such an action recommendation will help lead to faster convergence in reinforcement learning, and it is more effective than the LNNs-Shielding since it also has the positive recommendation function.

Let $\phi(s_t), lower_{n_i,s_t}, upper_{n_i,s_t}, ctrd(a_t,s_t)$ be the same as the definitions in the LNNs-Shielding method. The LNNs-Guide provides probabilities for recommendation of the action for each state input. The probability is calculated by

\begin{align}
P_{LNN} (a|s_t) &= \frac{e^{v(a,s_t)}}{\sum_{a_j \in A}e^{v(a_j,s_t)}}, \\
v(a,s_t) &= \frac{lower_{a,s_t} + upper_{a,s_t}}{2} - ctrd(a,s_t),
\end{align}

\noindent where $a$ is a targeted action for calculating the probability, and $A$ is all actions. Value~$v(a,s_t)$ represents the level of truth values for the propositions while discounting the contradiction value.

In reinforcement learning, the policy follows the probabilities calculated with the epsilon greedy algorithm in LSTM-DQN++~\cite{yuan2018counting}. In this work, we select the action~$a_t$ by

\begin{align}
a_t= \left\{\begin{array}{ll}
    {\rm arg max}_{a \in A} P_{LNN} (a|s_t) \: Q(s_t, a) &: \zeta \ge \epsilon \\
    {\rm random}_{a \in A} P_{LNN} (a|s_t)&: {\rm otherwise}
  \end{array}, \right.
\end{align}

\noindent where $\zeta$ is the current random value for epsilon greedy, and ${\rm random}_{a \in A} P$ takes an action in accordance with probabilities~$P$. In this equation, an action is selected on the basis of q-value and action probabilities from LNNs-Guide. Training steps by the reward signal are performed in the same way as in the reinforcement learning method.

\section{Evaluation}

\subsection{Experiments}
We evaluated the performance of the proposed method through experiments conducted in the TextWorld environment~\cite{cote2018textworld}. Our target was the Coin-collector game. We built LNNs that has useful knowledge based on external data. We fit the prepared data to the following rules and then trained the LNNs. 

\begin{itemize}
  \item $\lnot$ ``\texttt{visited all connected rooms}'' $\land$ ``\texttt{no coin in east room}''~$\Rightarrow$~$\lnot$ ``\texttt{go east}''
  \item ``\texttt{found east room}''~$\Rightarrow$~``\texttt{go east}''
  \item $\lnot$ ``\texttt{visited all connected rooms}'' $\land$ ``\texttt{no coin in west room}''~$\Rightarrow$~$\lnot$ ``\texttt{go west}''
  \item ``\texttt{found west room}''~$\Rightarrow$~``\texttt{go west}''
  \item $\lnot$ ``\texttt{visited all connected rooms}'' $\land$ ``\texttt{no coin in south room}''~$\Rightarrow$~$\lnot$ ``\texttt{go south}''
  \item ``\texttt{found south room}''~$\Rightarrow$~``\texttt{go south}''
  \item $\lnot$ ``\texttt{visited all connected rooms}'' $\land$ ``\texttt{no coin in north room}''~$\Rightarrow$~$\lnot$ ``\texttt{go north}''
  \item ``\texttt{found north room}''~$\Rightarrow$~``\texttt{go north}''
  \item ``\texttt{found coin in the room}''~$\Rightarrow$~``\texttt{take coin}''
\end{itemize}

\noindent The reason we have $\lnot$ ``\texttt{visited all connected rooms}'' as a rule is that the agent might have any preferred action at the dead-end of a path. The agent can go back to the visited room with this proposition when it is at the dead-end. 
%The proposition of ``\texttt{visited all connected rooms}'' is a result of AND operation for all value of ``\texttt{no coin in ? room}'' (\texttt{?} is filled a direction for the connected room). 
We prepared the LSTM-DQN++~\cite{yuan2018counting} method as a baseline method and tested it along with the proposed methods (LNNs-Shielding, LNNs-Guide). We set $\alpha=1$ for LNNs-Shielding.

\begin{figure}[t]
\begin{center}
\includegraphics[width=8cm]{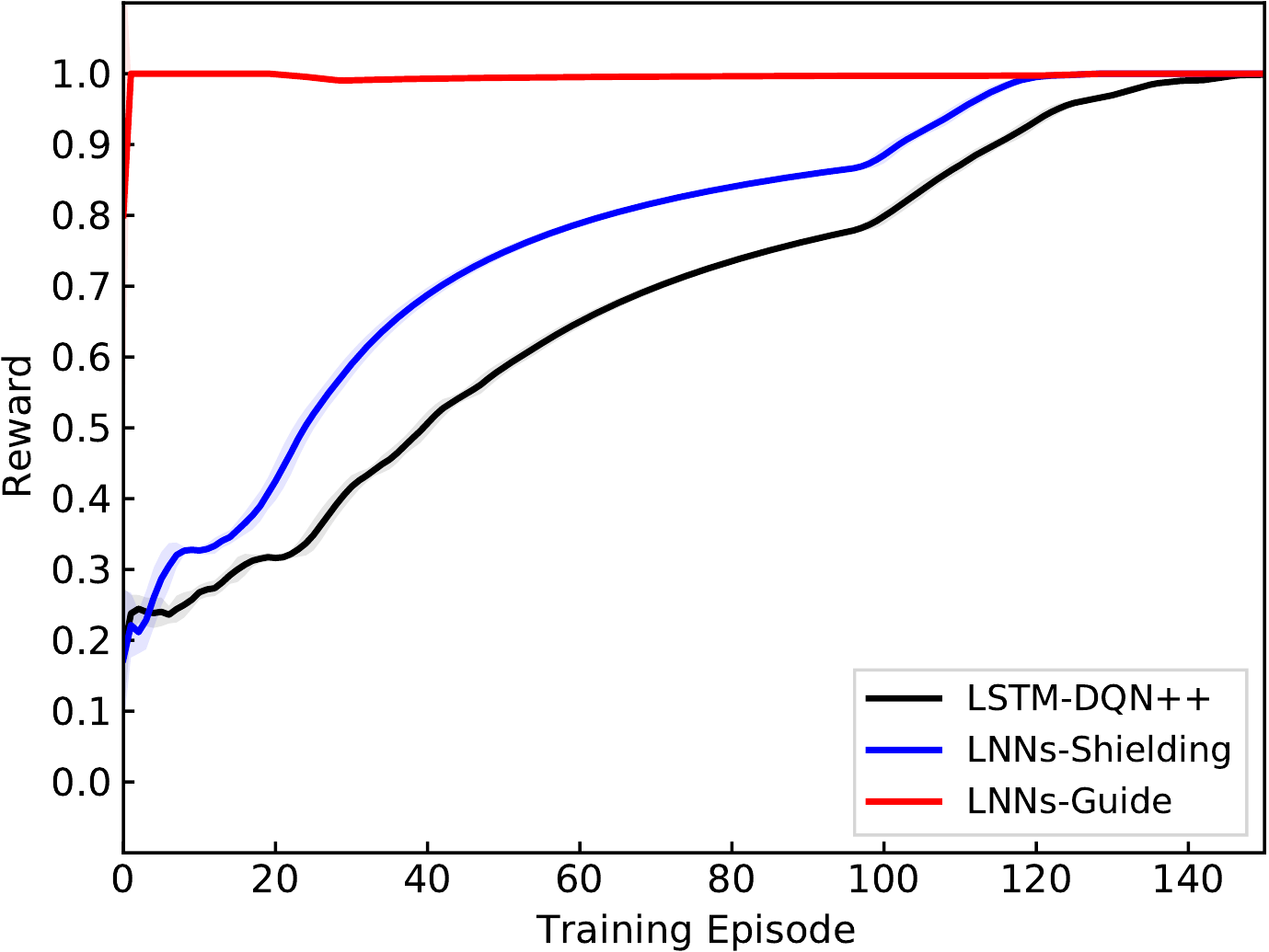}
\end{center}
\caption{Reward [0-1] curves for proposed and baseline methods with moving average ($N=5$). Shaded area represents the standard deviation value.}
\label{fig:reward_curve}
\end{figure}

\subsection{Results}

Figure~\ref{fig:reward_curve} shows the reward curves from the proposed and baseline methods. The LNNs-Shielding avoided previously visited rooms thanks to the rules in LNNs, such as ``\texttt{visited all connected rooms}'' $\land$ ``\texttt{no coin in east room}''~$\Rightarrow$~$\lnot$ ``\texttt{go east}'', so it had a better learning convergence than the baseline method. The LNNs-Guide converged the training extremely fast thanks to the effect of the positive recommendation, such as ``\texttt{found east room}''~$\Rightarrow$~``\texttt{go east}''. The reason LNNs-Shielding was weaker than LNNs-Guide is that, while LNNs-Shielding only prohibited an action when the given action had a high contradiction value, the LNNs-Guide provided action recommendations at every time step. This result is in line with our expectation and leads us to conclude that LNNs-Guide is the superior method for utilizing external knowledge. However, we also feel that LNNs-Guide may produce weaker results due to incorrect or fuzzy rules in the LNNs. To alleviate such concerns, we believe the interpretability of the rules in LNNs, which is the key benefit of LNNs, would be helpful to confirm the correctness of the trained rules. 

\section{Conclusion}

In this work, we have proposed a method that utilizes external knowledge represented in trainable logical neural networks and demonstrated through experiments that is has better convergence compared to a baseline method. For future work, we plan to apply this method to other complex games and train the policy directly in logical neural networks.

\bibliographystyle{named}
\bibliography{main}

\end{document}